\documentclass[num-refs, hidelinks]{nbdt-article}

\usepackage{siunitx}
\usepackage{float}
\usepackage[T1]{fontenc}    
\usepackage{hyperref}       
\usepackage{url}            
\usepackage{booktabs}       
\usepackage{amsfonts}       
\usepackage{nicefrac}       
\usepackage{microtype}      
\usepackage{xcolor}         
\usepackage{amsmath} 
\usepackage{amssymb}  
\usepackage{graphicx}
\usepackage{subfigure}
\usepackage{caption}
\usepackage{adjustbox}
\usepackage{algorithm}
\usepackage{algpseudocode}
\usepackage{algpseudocode}
\papertype{Original Article}

\title{Deep Direct Discriminative Decoders for High-dimensional Time-series Data Analysis}

\author[1,2,5]{Mohammad R. Rezaei}
\author[2,5]{Milos R. Popovic}
\author[6]{Uri T. Eden}
\author[1,2,3,5]{Milad Lankarany}
\author[4,\authfn{1}]{Ali Yousefi}

\affil[1]{Krembil Brain Institute, University Health Network, Toronto, Canada}
\affil[2]{KITE Research Institute,  Rehabilitation Institute, University Health Network, Toronto, Canada.}
\affil[3]{Department of Physiology, University of Toronto, Toronto, Canada.}
\affil[4]{Worcester Polytechnic Institute (WPI), Worcester, MA, USA.}
\affil[5]{Institute of Biomedical Engineering University of Toronto, Toronto, Canada.}
\affil[6]{Department of Mathematics and Statistics, Boston University, Boston, MA, USA.}

\corraddress{Ali Yousefi, Department of Computer Science at Worcester Polytechnic Institute, Worcester, MA, USA}
\corremail{ayousefi@wpi.edu}

\runningauthor{Rezaei, MR}

\begin{document}

\maketitle
\begin{abstract}
The state-space models (SSMs) are widely utilized in the analysis of time-series data. SSMs rely on an explicit definition of the state and observation processes. Characterizing these processes is not always easy and becomes a modeling challenge when the dimension of observed data grows or the observed data distribution deviates from the normal distribution.
Here, we propose a new formulation of SSM for high-dimensional observation processes. We call this solution the deep direct discriminative decoder (D4). The D4 brings deep neural networks' expressiveness and scalability to the SSM formulation letting us build a novel solution that efficiently estimates the underlying state processes through high-dimensional observation signal. We demonstrate the D4 solutions in simulated and real data such as Lorenz attractors, Langevin dynamics, random walk dynamics, and rat hippocampus spiking neural data and show that the D4 performs better than traditional SSMs and RNNs. The D4 can be applied to a broader class of time-series data where the connection between high-dimensional observation and the underlying latent process is hard to characterize.
\end{abstract}


\justifying

\section{Introduction}
The state-space model (SSM) is one of the well-established dynamical latent variable modeling frameworks successfully applied in the analysis of dynamical time-series data \cite{paninski2010new}. The Kalman filter, the most widely known form of SSMs, has been frequently utilized in characterizing a wide range of time-series data from healthcare \cite{zhang2015task}, navigation \cite{grewal1990application}, machine vision \cite{kiriy2002three}, and neuroscience \cite{eden2004dynamic}. The standard SSM consists of a state process that defines how the state, i.e. dynamical latent variables, evolves in time, and an observation process that defines the conditional distribution of the observed signals given the state variable(s) \cite{durbin2012time}. A modeling challenge in SSMs is to build an accurate conditional probability distribution of the observed signal given the state, specifically, for the cases where the dimension of the observed signal is high or it has a heavy-tailed distribution\cite{zoltowski2020general,linderman2017bayesian,glaser2020machine, batty2019behavenet}. In practice, the observation distribution model is simplified with assumptions such as conditional independence between dimensions of the signal or normal distribution for the observed signal. These assumptions might not hold in many datasets, including our research domain where we have neural activity of thousands of neurons\cite{steinmetz2021neuropixels, musick2009three}.\\
Recently, dynamical discriminative models such as recurrent neural networks (RNNs) and deep Kalman filter (Deep-KF) have been introduced for analyzing high-dimensional time series \cite{ krishnan2015deep, rezaei2018comparison, glaser2020recurrent}. These models do not face the same challenges as SSMs in characterizing high-dimensional data. RNNs and Deep-KF show a higher level of flexibility in characterizing complex distributions of observed signals as they take advantage of deep neural networks and non-linear models. On the other hand, characterizing high-dimensional data is challenging using SMMs since they typically rely on linear transformations to describe the data. However, RNNs and Deep-KF have their own modeling challenges \cite{gao2016linear,johnson2016structured, krishnan2015deep, archer2015black}. They primarily ignore the temporal dynamics and continuity of the underlying biological or physical systems in characterizing the observed dynamics, which potentially obscures the interpretability of inferred states. Additionally, their expressive power cannot be efficiently utilized given the limited size of data available in most neuroscience experiments\cite{krishnan2015deep,rangapuram2018deep}. 
Most recent modeling frameworks such as the discriminative Kalman filters (DKFs)\cite{burkhart2020discriminative} and direct discriminative decoders (DDD) \cite{rezaei2022direct} are designed to bring attributes of discriminative models, such as expressiveness and scalability, to the SSM framework. A critical advantage of SSMs is their explicit definition of the state dynamics, which allows for an optimal combination of information carried by the state and observation to infer the underlying latent dynamics. Thus, it is expected that the DKF and DDD solutions leverage the advantages of both SSM and discriminative models in characterizing observed signals and inferring the underlying dynamics. In these models, the observation process of SSMs is replaced by a discriminative model, thus avoiding the challenge of building a conditional probability distribution of the observed signal. Despite this advancement, a modeling concern with both DDD and DKF is that their proposed solution only addresses linear discriminative models with additive Gaussian noise and discards the history of the observed signals in their estimation of the state. In other words, the expressiveness of the discriminative model is not efficiently utilized, as DKF and DDD solutions do not address how DNN and RNN models can be embedded in their framework.\\
We propose deep direct discriminative decoder models (D4) and show how to attain DNN’s expressiveness as it optimally combines the current and history of the observation in its estimation of the underlying state. We show that the D4 will reach a high level of accuracy in the estimation of state dynamics and it provides a good generalization performance despite the limited data size. We tested D4 on simulation and real datasets, where the D4 outperformed traditional DNN, Deep-KF, DKF, and DDD models in multiple performance metrics including squared error (MSE), correlation coefficient (CC)(e.g., see Figure\ref{fig:D4_Lorenz}). The D4 can be applied to different modalities of time-series data without constraints on the modality or distribution of observed signals. In summary, we believe that the D4 solution facilitates the analysis of high-dimensional data where the connection between the high-dimensional observation and the underlying state process needs to attain a balanced level of interpretability and prediction accuracy.
\subsection{Backgrounds}
Consider the problem of modeling time-series data $\boldsymbol{y}_{1:K}, \boldsymbol{y}_k\in\mathbf{R}^N$, where $k=1,..., K$, using dynamical latent variables $\boldsymbol{x}_{1:K}, \boldsymbol{x}_k\in\mathbf{R}^M$ with a Markovian property. Under the SSM modeling framework, the joint probability distribution of latent variables and observations can be factorized by conditional probabilities of a generative process defined by
\begin{equation}
    p(\boldsymbol{x}_{1:K},\boldsymbol{y}_{1:K})
    = p(\boldsymbol{x}_{1})
    p(\boldsymbol{y}_{1}|\boldsymbol{x}_{1})
    \prod_{k=2}^{K}p(\boldsymbol{x}_{k}|\boldsymbol{x}_{k-1}) 
    p(\boldsymbol{y}_{k}|\boldsymbol{x}_{k}).
\end{equation}
The posterior distribution of $\boldsymbol{x}_{1:k}$ given $\boldsymbol{y}_{1:k}$, is defined by the following recursive solution
\begin{equation}
	\label{Eq-Bayesian-filtering}
    p(\boldsymbol{x}_{1:k}| \boldsymbol{y}_{1:k})\propto p(\boldsymbol{y}_{k}| \boldsymbol{x}_{k}) p(\boldsymbol{x}_{k}| \boldsymbol{x}_{k-1}) p(\boldsymbol{x}_{1:k-1}| \boldsymbol{y}_{1:k-1}),
\end{equation}
where the first term is the likelihood function given the current observation, the second term is the state process conditional distribution, and the last term is the posterior from the previous time index \cite{chen2003bayesian}.
A modeling challenge in SSM is to build the conditional probability of the observed signal. In practice, the likelihood function is simplified by an assumption like observations are conditionally independent given the state or they follow normal distributions. These assumptions are prohibitive in characterizing datasets such as neural data, where the observed signals include the spiking activity which can not be properly characterized by a normal distribution. These simplifications will potentially induce bias in the estimation of the underlying state process, causing an inaccurate inference of the state variables. The SSM modeling structure, specifically its generative model for the observation process, will avoid the utilization of many powerful tools such as DNNs in its characterization of observed signals and estimation of the underlying states. With this in mind, we propose a new variant of SMMs called the deep direct discriminative decoder (D4) model; a new modeling framework that fuses the advantages of DNN in the SSM to build a scalable and potentially accurate solution for characterizing high-dimensional dynamical time-series data.
\section{D4 model}
\begin{figure}[]
		\centering
		\includegraphics[width = .8\columnwidth]{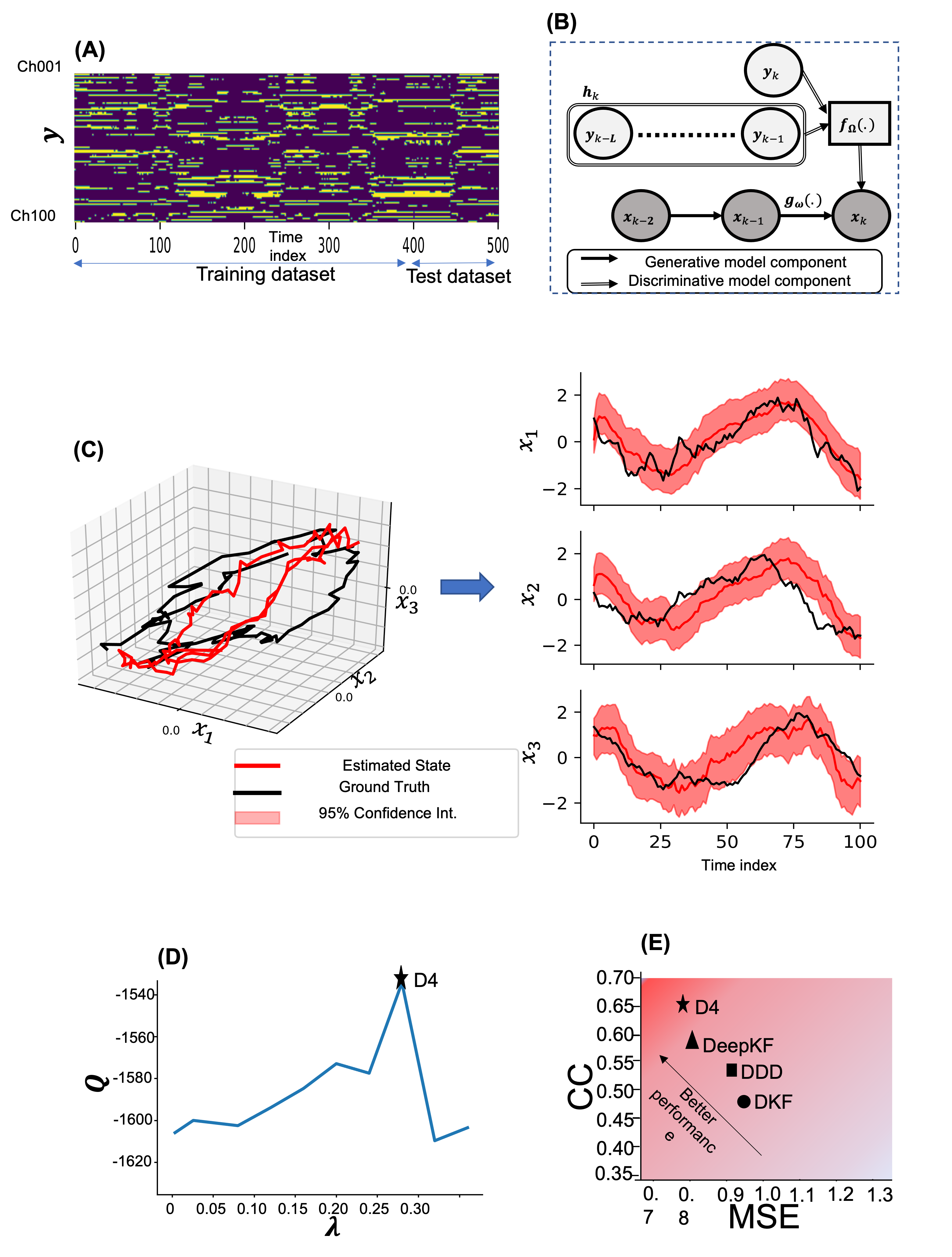}
		\caption{ \centering The D4 model application in a simulated high-dimensional dataset section  \ref{Lorenz-section}. A) Observation data include 100 simulated spiking channels using a point process intensity model from a multi-dimensional state trajectory. The length of the observation is 500 points, and the underlying state is defined by a Lorenz system. Further explanation of the simulation is discussed in Appendix \ref{app_point_process_obs}. B) Graphical representation of the D4 model. $\boldsymbol{g}_{\omega}(.)$ is the state transition process with a set of parameters defined by $\omega$ and $\boldsymbol{f}_{\Omega}(.)$ is a discriminative process with the parameter set defined by $\Omega$. C) The inferred trajectory using D4. The D4 successfully inferred the latent state trajectory as most of the points in the trajectory are covered by the 95\% HPD of the D4's predictions. D) Optimal $\lambda$, which controls the length of the history term in D4's discriminative models. See the section \ref{Lorenz-section} for a detailed explanation $\lambda$. E) The D4's inference performance compared with DDD\cite{rezaei2022direct}, DKF\cite{burkhart2020discriminative}, and Deep-KF\cite{krishnan2015deep}(see section \ref{Lorenz-section} for more details).  A more detailed correlation matrix comparison of inferred and true latent variables dynamics between D4 and Deep-KF models is shown in Figure \ref{fig:D4_DeepKF_cc}
 }.
		\label{fig:D4_Lorenz}
	\end{figure}
Let's assume $\boldsymbol{h}_k=\boldsymbol{y}_{1:k-1}$, we can rewrite the posterior distribution of $\boldsymbol{x}_{1:k}$ given $\boldsymbol{y}_{1:k}$ as
\begin{equation}
\label{eq:DDD1}
    p(\boldsymbol{x}_{1:k}|\boldsymbol{y}_{k},\boldsymbol{h}_{k})
    = \frac{p(\boldsymbol{x}_{1:k},\boldsymbol{y}_{k},\boldsymbol{h}_{k})}{p(\boldsymbol{y}_{k},\boldsymbol{h}_{k})}
    = \frac{p(\boldsymbol{y}_{k}|\boldsymbol{x}_{k},\boldsymbol{x}_{1:k-1},\boldsymbol{h}_{k})
    p(\boldsymbol{x}_{k},\boldsymbol{x}_{1:k-1},\boldsymbol{h}_{k})}{p(\boldsymbol{y}_{k},\boldsymbol{h}_{k})}.
\end{equation}
With the Markovian assumption of the state process and factorization rule, we can rewrite equation \ref{eq:DDD1} as
\begin{equation*}
    p(\boldsymbol{x}_{1:k}|\boldsymbol{y}_{k},\boldsymbol{h}_{k})=
\end{equation*}
\begin{equation}
\label{eq:DDD2}
     \frac{p(\boldsymbol{y}_{k}|\boldsymbol{x}_{k},\boldsymbol{h}_{k})
    p(\boldsymbol{x}_{k},\boldsymbol{x}_{1:k-1},\boldsymbol{h}_{k})}{p(\boldsymbol{y}_{k},\boldsymbol{h}_{k})}
    = \frac{p(\boldsymbol{y}_{k}|\boldsymbol{x}_{k},\boldsymbol{h}_{k})
    p(\boldsymbol{x}_{k}|\boldsymbol{x}_{k-1})
    p(\boldsymbol{x}_{1:k-1}|\boldsymbol{h}_{k})
    p(\boldsymbol{h}_{k})
    }{p(\boldsymbol{y}_{k},\boldsymbol{h}_{k})},
\end{equation}
where we replace $p(\boldsymbol{x}_k|\boldsymbol{x}_{1:k-1}, \boldsymbol{h}_k)$ with $p(\boldsymbol{x}_k|\boldsymbol{x}_{k-1})$. 
 A big assumption in traditional dynamical models such as KFs, is that the observations are conditionally independent of each other  given the corresponding state values.
This is a modeling assumption that may not be true in many dynamical data such as the spiking activity of neural ensembles\cite{pillow2008spatio}. To overcome this, here we  propose a structure for the observation process that allows us to incorporate a history of previous observations and take advantage of state-of-the-art discriminative models such as DNNs as the observation model. To be able to incorporate a history of observations in the observation model, we need to reformulate the observation process to a discriminative one. We can use the Bayes rule to change $p(\boldsymbol{y}_{k}|\boldsymbol{x}_{k},\boldsymbol{h}_{k})$ and replacing $\boldsymbol{h}_k$ with $\{\boldsymbol{y}_{k-1},\boldsymbol{h}_{k-1}\}$, we can rewrite the posterior as
\begin{equation*}
    p(\boldsymbol{x}_{1:k}|\boldsymbol{y}_{k},\boldsymbol{h}_{k})=
     \frac{p(\boldsymbol{x}_{k}|\boldsymbol{y}_{k},\boldsymbol{h}_{k})
     p(\boldsymbol{y}_{k},\boldsymbol{h}_{k})
    p(\boldsymbol{x}_{k}|\boldsymbol{x}_{k-1})
    p(\boldsymbol{x}_{1:k-1}|\boldsymbol{h}_{k})
    p(\boldsymbol{h}_{k})}
    {p(\boldsymbol{y}_{k},\boldsymbol{h}_{k})
    p(\boldsymbol{x}_{k}|\boldsymbol{h}_{k})
    p(\boldsymbol{h}_{k})}=
\end{equation*}
\begin{equation}
    \label{eq:DDD3}
    \frac{
    p(\boldsymbol{x}_{k}|\boldsymbol{y}_{k},\boldsymbol{h}_{k})}{
    p(\boldsymbol{x}_{k}|\boldsymbol{h}_{k})
    } p(\boldsymbol{x}_{k}|\boldsymbol{x}_{k-1})
    p(\boldsymbol{x}_{1:k-1}|\boldsymbol{y}_{k-1},\boldsymbol{h}_{k-1})
\end{equation}
where $p(\boldsymbol{x}_{k}|\boldsymbol{x}_{k-1})$ is the state transition process and we call $p(\boldsymbol{x}_{k}|\boldsymbol{y}_{k},\boldsymbol{h}_{k})$ the prediction process. The $p(\boldsymbol{x}_{1:k-1}|\boldsymbol{y}_{k-1},\boldsymbol{h}_{k-1})$ becomes the posterior distribution from the previous time index and the denominator term, $p(\boldsymbol{x}_{k}|\boldsymbol{h}_{k})$, can be expanded by $p(\boldsymbol{x}_{k}|\boldsymbol{h}_{k})=\int d\boldsymbol{x}_{k-1} p(\boldsymbol{x}_{k}|\boldsymbol{x}_{k-1}) p(\boldsymbol{x}_{k-1}|\boldsymbol{h}_{k-1},\boldsymbol{y}_{k-1})$. Equation \ref{eq:DDD3} gives a recursive solution to calculate the posterior at time index $k$. In this formulation, the ratio term $
    p(\boldsymbol{x}_{k}|\boldsymbol{y}_{k},\boldsymbol{h}_{k})/
    p(\boldsymbol{x}_{k}|\boldsymbol{h}_{k})
    $ is equivalent to the update term in the SSM and represents the amount of information carried by the current observation. If the current observation is not informative, the ratio is close to one which corresponds to a flat likelihood in the SSM for the observed signal; on the other hand, it will push the one-step prediction distribution toward a new domain of states implied by the observed data.\\
As Figure\ref{fig:D4_Lorenz}.B shows, the D4 is comprised of two equations: a) a state transition equation, and b) a prediction process equation. The state transition equation at time index $k$ is defined by 
	\begin{equation}
	\label{eq_state_tran}
	\boldsymbol{x}_{k}|\boldsymbol{x}_{k-1} \sim g(\boldsymbol{x}_{k-1};\boldsymbol{\omega} ),
	\end{equation}
     where $\boldsymbol{\omega}$ is the model parameters of the conditional distribution. The prediction process is defined by 
	\begin{equation}
	\label{eq_prediction_process}
	\boldsymbol{x}_{k}|\boldsymbol{y}_{k},\boldsymbol{h}_{k} \sim f(\boldsymbol{y}_{k},\boldsymbol{h}_{k};\boldsymbol{\Omega} ).
	\end{equation}
	In practice, $\boldsymbol{h}_{k}$ is assumed to be a subset of the observation from previous time points, and $\boldsymbol{\Omega}$ is the model-free parameters of the prediction process defined by the discriminative function $f$.\\
Equations \ref{eq_state_tran} and \ref{eq_prediction_process} define the D4 model, where, the observation equation of SSM is replaced by a discriminative process. The discriminative process can be built using state-of-the-art discriminative models such DNNs, RNNs, CNNs, or variants of these models. The flexibility in picking the discriminative process allows us to benefit from the scalability and expressiveness of these models, where these choices make the D4 agnostic to the modality of the input data.\\ 
Here we defined the D4 model structure, equations \ref{eq_state_tran}, \ref{eq_prediction_process}, and \ref{eq:DDD3} give a recursive solution to compute the posterior distribution of the states variables. The denominator term in equation \ref{eq:DDD3} requires an integration, which adds up to the computational complexity of the filter estimation. In the following sections, we propose an efficient sampling solution that allows us to efficiently draw a sample of state trajectories recursively for the state posterior distribution, defined by $P(\boldsymbol{x}_{1:K}\mid \boldsymbol{y}_{1:K})$. We also discuss the computational complexity of the sampling solution for the posterior estimation in the D4 and compare it with the SSM one. We then discuss the training solution for the D4; we will see how the $P(\boldsymbol{x}_{1:K}\mid \boldsymbol{y}_{1:K})$ samples will be used in the training of the D4 components and discuss how in practice we can reduce the D4 computational cost by controlling the $\boldsymbol{h}_k$ term.
\subsection{Smoothed sequential importance sampling}
\label{sec-sampling}
Equation \ref{eq:DDD3} defines the posterior distribution of the state variables given the observation. Like sequential Monte Carlo sampling \cite{doucet2009tutorial}, we assume a proposal distribution define by $q_k(\boldsymbol{x}_k\mid \boldsymbol{x}_{k-1})$. 
Let's assume at time index $k$, we draw $D$ samples using the proposal distribution to give the samples from the previous time point. With this assumption, the weight for $d$th sample, $w_{k\mid k}^{(d)}$, is defined by 
\begin{equation*}
    \boldsymbol{x}^{(d)}_{k} \sim q(\boldsymbol{x}^{(d)}_{k}|\boldsymbol{x}^{(d)}_{k-1},\boldsymbol{y}_{k}),
\end{equation*}
\begin{equation}
\label{eq_sampling_w}
    w_{k\mid k}^{(d)} := \frac{
p(\boldsymbol{x}^{(d)}_{k}|\boldsymbol{x}^{(d)}_{k-1})    p(\boldsymbol{x}^{(d)}_{k}|\boldsymbol{y}_{k},\boldsymbol{h}_{k}) }{
    q(\boldsymbol{x}^{(d)}_{k}|\boldsymbol{x}^{(d)}_{k-1},\boldsymbol{y}_{k}) p(\boldsymbol{x}^{(d)}_{k}\mid \boldsymbol{h}_{k}) 
    }
\end{equation}
The second term in the denominator of equation \ref{eq_sampling_w} can be numerically calculated using the samples from the previous time index, $p(\boldsymbol{x}^{(d)}_{k}|\boldsymbol{h}_{k}) = \sum_{\boldsymbol{x}^{(d)}_{k-1}} p(\boldsymbol{x}^{(d)}_{k}\mid \boldsymbol{x}^{(d)}_{k-1})p(\boldsymbol{x}^{(d)}_{k-1}|\boldsymbol{h}_{k})$, or can be approximated using  the theory of Laplace approximations \cite{wong2001asymptotic}. By considering $g(\boldsymbol{x}_{k-1})=\log ( p(\boldsymbol{x}_{k-1}|\boldsymbol{h}_{k-1},\boldsymbol{y}_{k-1})p(\boldsymbol{x}_{k}|\boldsymbol{x}_{k-1})) $ and $\boldsymbol{x}^*_{k-1}=\arg \max_{\boldsymbol{x}_{k-1}} g(\boldsymbol{x}_{k-1}) $ and using the Laplace approximation we have 
$$
p(\boldsymbol{x}_{k}|\boldsymbol{h}_{k})=\int d\boldsymbol{x}_{k-1} \exp{(\log (g(\boldsymbol{x}_{k-1})))}\approx
$$
$$
\int d\boldsymbol{x}_{k-1} \exp{\left ( g(\boldsymbol{x}^*_{k-1}) +  \triangledown g(\boldsymbol{x}_{k-1})|_{\boldsymbol{x}^*_{k-1}}(\boldsymbol{x}_{k-1}-\boldsymbol{x}^*_{k-1})^T + \frac{1}{2}(\boldsymbol{x}_{k-1}-\boldsymbol{x}^*_{k-1})^T\triangledown^2 g(\boldsymbol{x}_{k-1})|_{\boldsymbol{x}^*_{k-1}}(\boldsymbol{x}_{k-1}-\boldsymbol{x}^*_{k-1}) \right )}=
$$
\begin{equation}
\exp{ g(\boldsymbol{x}^*_{k-1})} \int d\boldsymbol{x}_{k-1} \exp{\left (   \frac{1}{2}(\boldsymbol{x}_{k-1}-\boldsymbol{x}^*_{k-1})^T\triangledown^2 g(\boldsymbol{x}_{k-1})|_{\boldsymbol{x}^*_{k-1}}(\boldsymbol{x}_{k-1}-\boldsymbol{x}^*_{k-1}) \right )}
\end{equation}
Notice that here ${\triangledown g(\boldsymbol{x}_{k-1})|_{\boldsymbol{x}^*_{k-1}}}=0$ and $p(\boldsymbol{x}_{k}|\boldsymbol{h}_{k})$ approximated by a distribution proportional to a Gaussian density with mean  $\boldsymbol{x}^*_{k-1}$ and variance $-\{\triangledown^2 g(\boldsymbol{x}_{k-1})|_{\boldsymbol{x}^*_{k-1}}\}^{-1}$. We have a closed form for the integral as
\begin{equation}
p(\boldsymbol{x}_{k}|\boldsymbol{h}_{k})\approx\exp{ g(\boldsymbol{x}^*_{k-1})} {(2\pi)^{M/2}}{{|{\{\triangledown^2 g(\boldsymbol{x}_{k-1})|_{\boldsymbol{x}^*_{k-1}}}\}^{-1}|}^{1/2}}
\end{equation}
 Where $\boldsymbol{x}_k $ is an M-dimensional vector. Therefore Laplace's approximation converts the integration problem to a maximization problem where we need to find a maximizer, $\boldsymbol{x}^*_{k-1}$, for $g(\boldsymbol{x}_{k-1})$. Using maximizer $\boldsymbol{x}^*_{k-1}$ one can easily approximate the denominator of the equation \ref{eq_sampling_w} for each sample $d$ and avoid computationally complex integration. In the case of assuming Gaussian approximation for the state transition process and prediction process and considering that $\log$ is a monotonic function, $\boldsymbol{x}^*_{k-1}$ is the mean of the joint Gaussian distribution of both processes at time index $k-1$. If $\boldsymbol{x}^*_{k-1}$ or the derivatives of $g$ are not available in closed form, then the Laplace approximation can be slow owing to the need to solve an optimization problem at each time step.\\
 In cases where the state process is a diffusion process (a random walk), due to the fact that diffusion processes do not change the mode of distribution and only change the variance of the distribution, the maximizer for $g(\boldsymbol{x}_{k-1})$ is equal to the maximizer of $\boldsymbol{x}^*_{k-1}= \arg \max_{\boldsymbol{x}_{k-1}} \log ( p(\boldsymbol{x}_{k-1}|\boldsymbol{h}_{k})) $. \\
Using equation \ref{eq_sampling_w}, we draw samples from the filter estimation. To draw samples from the smoother estimate of the state, we use the forward filtering and backward smoothing (FFBS) formula suggested in \cite{kitagawa1996monte}.
FFBS first runs the filter solution define by equation \ref{eq_sampling_w} for the D4; it then reweighs particles with the backward recursion defined by
\begin{equation}
\label{eq_smoothing_w}
    w_{k\mid K}^{(d)} = \overline{w}_{k\mid k}^{(d)} \left [ \sum_{i=1}^{K} {w}_{k+1\mid K}^{(i)} \frac{p(\boldsymbol{x}^{(i)}_{k+1}|\boldsymbol{x}^{(k)}_{k})}{\sum_{j=1}^{K}  \overline{w}_{k\mid k}^{(j)} p(\boldsymbol{x}^{(i)}_{k+1}|\boldsymbol{x}^{(k)}_{j})} \right ]
\end{equation}
where $\overline{w}_{K\mid k}^{(d)} ={w}_{K\mid k}^{(d)}/ \sum_{j=1}^{D} {w}_{K\mid k}^{(j)} $ with $w_{K\mid K}^{(d)} = \overline{w}_{K\mid K}^{(d)}$. Using equations \ref{eq_sampling_w} and \ref{eq_smoothing_w}, we can draw samples from the posterior of the state given the whole observation. Note that we will do the resampling at each time index of the filter solution. 
\subsection{ Computational complexity of SMMs vs D4}
The filtering for SSM, defined in equation \ref{Eq-Bayesian-filtering} for SSM requires $O(NK)$ operations to sample one trajectory of the state approximately distributed according to $p(\boldsymbol{x}_{1:K}|\boldsymbol{y}_{1:K})$\cite{doucet2009tutorial}.
On the other hand, the D4 requires $O(NK^2)$ operations to sample one path for the same distribution, if we consider $\boldsymbol{h}_k= \boldsymbol{y}_{1:k-1}$. In practice, we replace the $\boldsymbol{h}_k$ with a fixed length history $\boldsymbol{h'}_k=\{\boldsymbol{y}_j\}_{j=k-1}^{j=k-L-1}$ which can reduce the computational cost to $O(NKL)$. In the following sections, we discuss the training algorithm where we can estimate the model free parameters ($\omega,\  \Omega$) along with the optimal history length, as a result, the training algorithm outcome for the history length will impact the computational complexity of the filter and smoother solution.
\subsection{ D4 Model Training}
With the D4 training, we maximize the likelihood of the observation data by tuning the model-free parameters. The D4 free parameters include the state process parameters - e.g. $\omega$, and the discriminative model free parameters  - e.g $\Omega$. Note that the history term, $\boldsymbol{h}_k$, needs to be identified, which will change the optimal values of $\boldsymbol{\omega}$ and $\boldsymbol{\Omega}$ of the D4 model. Note that the state variables, $\boldsymbol{x}_k$, are not directly observed; thus, the training requires the estimation of the state process \cite{kenny1984estimating}. For this setting, we can use the EM algorithm, which let us find the ML estimates of the model parameters \cite{myung2003tutorial}. The EM alternates between computing the expected complete log-likelihood according to the posterior estimation of the state (the E step) and maximizing this expectation, $\boldsymbol{Q}$ function, by updating the model free parameters (the M step). The $\boldsymbol{Q}$ is defined by
	\begin{equation}
	\label{eq_lower_def}
	\boldsymbol{Q}(\boldsymbol{\theta}|\boldsymbol{\theta}^{r})={\mathbb{E}}_{{\boldsymbol{x}}_{{0:K}}|{\boldsymbol{y}}_{{1:K}};\boldsymbol{\theta}^{r}}
	[\log (p({\boldsymbol{x}}_{\boldsymbol{0}};\boldsymbol{\omega}_0) \prod^K_{k=1}p({\boldsymbol{x}}_{{k}}|{\boldsymbol{x}}_{{k-1}};\boldsymbol{\omega})
	\prod^K_{k=1}
	\frac{p({\boldsymbol{x}}_{{k}}|{\boldsymbol{y}}_{{k}},{\boldsymbol{h}}_{{k}};\boldsymbol{\Omega})}{ p({\boldsymbol{x}}_{{k}}|{\boldsymbol{h}}_{{k}};\boldsymbol{\Omega})
	})]
	\end{equation}
	where $\boldsymbol{\theta}^r = \{ \boldsymbol{\omega}_0^r, \boldsymbol{\Omega}^r,\boldsymbol{\omega}^r \}$ represents the model parameters estimated by maximizing $\boldsymbol{Q}$ at the previous iteration of the EM algorithm \cite{yousefi2015cognitive}. For the simplicity of notation, we use $\mathbb{E}_K$ for ${\mathbb{E}}_{{\boldsymbol{x}}_{{0:K}}|{\boldsymbol{y}}_{{1:K}};\boldsymbol{\theta}^{r}}$ in the rest of the paper.
 We can expand the $\boldsymbol{Q}$ function, as
	\begin{equation*}
	    \boldsymbol{Q}(\boldsymbol{\theta}|\boldsymbol{\theta}^{r})=\mathbb{E}_K
	[
	\log \; p({\boldsymbol{x}}_{{0}};\boldsymbol{\omega}_0) +
	\end{equation*}
	\begin{equation}
	\label{eq_Qvale_1}
	 \sum_{k=1}^{K} \log\; p({\boldsymbol{x}}_{{k}}|{\boldsymbol{x}}_{{k-1}};\boldsymbol{\omega}) +
	\sum_{k=1}^{K} \log \; p({\boldsymbol{x}}_{{k}}|{\boldsymbol{y}}_{{k}},{\boldsymbol{h}}_{{k}};\boldsymbol{\Omega})] -{\mathbb{E}}_K[
	\sum_{k=1}^{K} \log \; p({\boldsymbol{x}}_{{k}}|{\boldsymbol{h}}_{{k}};\boldsymbol{\Omega})
	]\end{equation}
For the cases where the state process is linear with additive Gaussian noise and the prediction process is a multi-variate Gaussian, we can find a closed-form solution for different terms of the $\boldsymbol{Q}$ function. However, for almost all other cases such as when the prediction process is a non-linear function of the observation or the state transition represents non-linear dynamics, it is hard to find a closed-form solution for $\boldsymbol{Q}$ function given the model free parameters. Instead, we can use the sampling technique, described in the section \ref{sec-sampling}, to calculate the expectation for the first term, $\mathbb{E}_K[.]$, of equation \ref{eq_Qvale_1}; however, finding the expectation for the last term is still challenging given it involves a second integration over the state variable. To address this challenge, we show that we can find an approximation for the last term which turns into a lower bound for the $\boldsymbol{Q}$. We can write the last term in $\boldsymbol{Q}$ function as
	\begin{equation*}
	  {\mathbb{E}}_K[
	\sum_{k=1}^{K} \log \; p({\boldsymbol{x}}_{{k}}|{\boldsymbol{h}}_{{k}};\boldsymbol{\Omega})
	]= \sum_{k=1}^{K} \int d\boldsymbol{x}_{k} p(\boldsymbol{x}_{k}|\boldsymbol{y}_{1:K},\boldsymbol{\theta}^{r})\log p({\boldsymbol{x}}_{{k}}|{\boldsymbol{h}}_{{k}};\boldsymbol{\Omega})=
	\end{equation*}
	\begin{equation*}
	    =\sum_{k=1}^{K}\int  p(\boldsymbol{x}_{k}|\boldsymbol{y}_{1:K},\boldsymbol{\theta}^{r})\log \frac{p({\boldsymbol{x}}_{{k}}|{\boldsymbol{h}}_{{k}};\boldsymbol{\Omega}) p(\boldsymbol{x}_{k}|\boldsymbol{y}_{1:K},\boldsymbol{\theta}^{r})}{p(\boldsymbol{x}_{k}|\boldsymbol{y}_{1:K},\boldsymbol{\theta}^{r})} d\boldsymbol{x}_{k}
	\end{equation*}
	\begin{equation}
	  = \sum_{k=1}^{K} (\int d\boldsymbol{x}_{k} p(\boldsymbol{x}_{k}|\boldsymbol{y}_{1:K},\boldsymbol{\theta}^{r})\log \frac{p({\boldsymbol{x}}_{{k}}|{\boldsymbol{h}}_{{k}};\boldsymbol{\Omega}) }{p(\boldsymbol{x}_{k}|\boldsymbol{y}_{1:K},\boldsymbol{\theta}^{r})} +\int d\boldsymbol{x}_{k} p(\boldsymbol{x}_{k}|\boldsymbol{y}_{1:K},\boldsymbol{\theta}^{r}) \log p(\boldsymbol{x}_{k}|\boldsymbol{y}_{1:K},\boldsymbol{\theta}^{r}),
	\end{equation}
	where the first term here is the negative of KL divergence between $p(\boldsymbol{x}_{k}|\boldsymbol{y}_{1:K},\boldsymbol{\theta}^{r})$ and $p({\boldsymbol{x}}_{{k}}|{\boldsymbol{h}}_{{k}};\boldsymbol{\Omega})$, and the second term is the negative of the entropy of the ${\boldsymbol{x}}_{{k}}$ posterior distribution given the whole observation. Notice that $\mathbb{H}(p(\boldsymbol{x}_{k}|\boldsymbol{y}_{1:K},\boldsymbol{\theta}^{r})$ is independent of the free parameters $\boldsymbol{\theta}$.
	 As a result, we can rewrite $\boldsymbol{Q}$ as
	\begin{equation*}
	\boldsymbol{Q}(\boldsymbol{\theta}|\boldsymbol{\theta}^{r})=\mathbb{E}_K
	[
	\log \; p({\boldsymbol{x}}_{{0}};\boldsymbol{\omega}_0) + \sum_{k=1}^{K} \log\; p({\boldsymbol{x}}_{{k}}|{\boldsymbol{x}}_{{k-1}};\boldsymbol{\omega}) +
	\end{equation*}
	\begin{equation}
	\label{eq:Qd}
	\sum_{k=1}^{K} \log \; p({\boldsymbol{x}}_{{k}}|{\boldsymbol{y}}_{{k}},{\boldsymbol{h}}_{{k}};\boldsymbol{\Omega})]+
	    \sum_{k=1}^{K} \mathbb{KL}(p(\boldsymbol{x}_{k}|\boldsymbol{y}_{1:K},\boldsymbol{\theta}^{r}) \parallel p({\boldsymbol{x}}_{{k}}|{\boldsymbol{h}}_{{k}};\boldsymbol{\Omega}))+ \mathbb{H}(p(\boldsymbol{x}_{k}|\boldsymbol{y}_{1:K},\boldsymbol{\theta}^{r})
	\end{equation}
	The KL term in equation\ref{eq:Qd} is positive; thus, a lower bound for the $\boldsymbol{Q}$ can be defined as the expectation of the likelihood terms defined by the state transition and prediction processes. The KL term can not be zero; however, it can generate smaller values as the length of the history term in the discriminative model grows.\\ 
	Note that the history term has two contrasting effects in the $\boldsymbol{Q}$ function. As the history length grows, the $\boldsymbol{Q}$ value grows as the prediction process, the $\log p({\boldsymbol{x}}_{{k}}|{\boldsymbol{y}}_{{k}},{\boldsymbol{h}}_{{k}};\boldsymbol{\Omega})$, will better fit the state, where it brings the value of the KL down. Thus, we can assume the KL term in equation \ref{eq:Qd}, penalizes the length of $\boldsymbol{h}_{k}$ and prevents unconstrained growth of the $\boldsymbol{h}_{k}$. This dynamics of KL suggests that there is an optimal history length that not only contributes to a better prediction of the state but also pushes $\boldsymbol{Q}$ to a higher value. We assume that the sum of KL and H over the whole time indices should be smaller than a pre-set threshold $\epsilon$. With this assumption, we can approach our model training as a regularized maximum likelihood problem defined by
\begin{equation*}
 \max_{\boldsymbol{\theta}}\mathbb{E}_K
	[\log \; p({\boldsymbol{x}}_{{0}};\boldsymbol{\omega}_0) + \sum_{k=1}^{K} \log\; p({\boldsymbol{x}}_{{k}}|{\boldsymbol{x}}_{{k-1}};\boldsymbol{\omega}) + \sum_{k=1}^{K} \log \; p({\boldsymbol{x}}_{{k}}|{\boldsymbol{y}}_{{k}},{\boldsymbol{h}}_{{k}};\boldsymbol{\Omega})]
\end{equation*}
\begin{equation}
	\label{eq-objective-pre-final}
	s.t. \ \ 
	\sum_{k=1}^{K}\mathbb{KL}(p(\boldsymbol{x}_{k}|\boldsymbol{y}_{1:K},\boldsymbol{\theta}^{r}) \parallel p({\boldsymbol{x}}_{{k}}|{\boldsymbol{h}}_{{k}};\boldsymbol{\Omega}))+\mathbb{H}(p(\boldsymbol{x}_{k}|\boldsymbol{y}_{1:K},\boldsymbol{\theta}^{r}) )<\epsilon
\end{equation}
Re-writing equation \ref{eq-objective-pre-final} as a Lagrangian under the KKT conditions \cite{kuhn1951w, wan2001dual}, we obtain
\begin{equation*}
 \boldsymbol{Q}(\boldsymbol{\theta}, \lambda, \mid \boldsymbol{\theta}^{r}) = \mathbb{E}_K
	[\log \; p({\boldsymbol{x}}_{{0}};\boldsymbol{\omega}_0) + \sum_{k=1}^{K} \log\; p({\boldsymbol{x}}_{{k}}|{\boldsymbol{x}}_{{k-1}};\boldsymbol{\omega}) + \sum_{k=1}^{K} \log \; p({\boldsymbol{x}}_{{k}}|{\boldsymbol{y}}_{{k}},{\boldsymbol{h}}_{{k}};\boldsymbol{\Omega})]
\end{equation*}
\begin{equation}
	\label{eq-objective-final}
	-\lambda( 
	\sum_{k=1}^{K} \mathbb{KL}(p(\boldsymbol{x}_{k}|\boldsymbol{y}_{1:K},\boldsymbol{\theta}^{r}) \parallel p({\boldsymbol{x}}_{{k}}|{\boldsymbol{h}}_{{k}};\boldsymbol{\Omega}))+\mathbb{H}(p(\boldsymbol{x}_{k}|\boldsymbol{y}_{1:K},\boldsymbol{\theta}^{r}))
\end{equation}
$\lambda\geq 0$ keeps the KL non-zero by putting pressure to shrink history length. 
\begin{algorithm}[t]
\small
\caption{D4 Learning Algorithm} 
\label{alg_D4_learning}
\begin{algorithmic}[1]
    \Procedure{Regularized-EM-for-D4}{$\boldsymbol{y}_{1:K}, \boldsymbol{\theta}^{(0)}, \lambda, D, L$}
        \State $ \boldsymbol{h}_{k}\xleftarrow{}\{\boldsymbol{y}_{k-L:k-1} \}, \boldsymbol{Q}^{0}\leftarrow{} 0$
        \State \textbf{Do}
            \State $\ \ \ \ \ \boldsymbol{Q}^{max}\leftarrow{}\boldsymbol{Q}^r$
            \State $\ \ \ \ \ \boldsymbol{\tilde{x}}^{1:D}_{1:K}\leftarrow{}\boldsymbol{x}^{1:D}_{1:K}$ 
            \State $\ \ \ \ \ \ $ Sample D smoothed trajectories using equations \ref{eq_sampling_w} and \ref{eq_smoothing_w}, $ \boldsymbol{x}^{1:D}_{1:K} \sim p_{\boldsymbol{\theta}}(\boldsymbol{x}^{1:D}_{1:K}\mid \boldsymbol{y}_{1:K})$
            \State $\ \ \ \ \ \boldsymbol{\theta}^{r}, \boldsymbol{Q}^{r} = Update-Model(\boldsymbol{y}_{1:K},\boldsymbol{\tilde{x}}^{1:D}_{1:K},\boldsymbol{x}^{1:D}_{1:K},\boldsymbol{\theta}^{(r-1)}, \lambda)$
        \State \textbf{DoWhile}$\{\boldsymbol{Q}^{r}>\boldsymbol{Q}^{max}\}$
        \State \Return $\boldsymbol{\theta}^{(r-1)}, \boldsymbol{Q}^{max}$
        \EndProcedure
\Procedure{Update-Model}{$\boldsymbol{y}_{1:K},\boldsymbol{\tilde{x}}^{1:D}_{1:K},\boldsymbol{x}^{1:D}_{1:K},\boldsymbol{\theta}^{(r-1)}, \lambda$}  
    \State $\{\boldsymbol{\omega}_0^{(r-1)},\boldsymbol{\omega}^{(r-1)},\boldsymbol{\Omega}^{(r-1)}\} = \boldsymbol{\theta}^{(r-1)}$
    \State Update $\boldsymbol{Q}^{r}$ using equation \ref{eq-objective-final} evaluate at $\boldsymbol{\theta}^{(r-1)}$
    \State  Update $\boldsymbol{\Omega}^{r}$, $\boldsymbol{\omega}^{r}$, and $\boldsymbol{\omega}_0^{r}$  using gradients calculated by equations \ref{eq_lower_bound_grad_f}, \ref{app_eq_grad_omega}, and \ref{app_eq_grad_omega0}; respectively
     \State \Return $\{\boldsymbol{\omega}_0^{r}, \boldsymbol{\omega}^{r},\boldsymbol{\Omega}^{r}\} , \boldsymbol{Q}^{r}$
     \EndProcedure
\end{algorithmic}
\end{algorithm}
\subsection{Stochastic optimization for parameter estimation}
We can optimize the objective function defined in equation \ref{eq-objective-final} by using a stochastic optimization algorithm. Stochastic optimization algorithms follow noisy gradients to reach the optimum of an objective function. The gradient with respect to the state transition parameters is already derived in \cite{kingma2019introduction}, which can be found in equations \ref{app_eq_grad_omega0} and \ref{app_eq_grad_omega} of Appendix \ref{app_gradients}. We also require to find the gradient of $\boldsymbol{Q}$ with respect to $\boldsymbol{\Omega}$. This gradient is defined by 
	     \begin{equation}
	     \label{eq_lower_bound_grad_f}
	          \triangledown_{\boldsymbol{\Omega}} \boldsymbol{Q}(\boldsymbol{\theta}|\boldsymbol{\theta}^{r}) \simeq \frac{1}{D} \sum_{d=1}^{D}\sum_{k=1}^{K} \triangledown_{\boldsymbol{\Omega}} \log p({\boldsymbol{\hat{x}}}^{(d)}_{{k}}|{\boldsymbol{y}}_{{k}},{\boldsymbol{h}}_{{k}};\boldsymbol{\Omega}) + \lambda \triangledown_{\boldsymbol{\Omega}} \log p({\boldsymbol{\hat{x}}}^{(d)}_{{k}}|{\boldsymbol{h}}_{{k}};\boldsymbol{\Omega}),
	     \end{equation}
where $\{\boldsymbol{\hat{x}}^{(d)}_{k}\}_{k=1}^{k=K}$ is $d$th state trajectory path derived from the smoothed posterior with parameter set $\boldsymbol{\theta}^{r}$, see more details in Appendix \ref{app_gradients}. Given equation\ref{eq_lower_bound_grad_f}, we can optimize the $\boldsymbol{\Omega}$ using a stochastic back-propagation similar to the ones proposed for variational-autoencoder models \cite{kingma2013auto}.\\
Algorithm \ref{alg_D4_learning} presents the learning steps for the D4 using stochastic optimization. In developing the learning algorithm, we assumed that $\boldsymbol{x}_{k}$ is unobserved, when $\boldsymbol{x}_{k}$ is known, the learning procedure follows the same steps, whilst the expectation, $\mathbb{E}_K$, is replaced by the state values constructing the trajectory of the state.
\section{Experiments}
We demonstrate the utility of the D4 by testing its different modeling steps on simulated data created for non-linear dynamical systems with high-dimensional observations.
The simulated experiments are 1) Langevin dynamics with point-process observations, 2) Lorenz attractor with point-process observations, and 3) 1-D random walk model with a non-linear mapping onto 20-dimensional space as the observed signal (results for this example are shown Appendix \ref{app-simulaion-1D}). We selected these experiments to cover diverse non-linear dynamical systems with physically meaningful state variables and different types of observations and noise processes. We show that the D4 accurately estimates the non-linear latent dynamics. We also apply the D4 on a decoding problem using the rat hippocampus data while the rat forages a W-shaped maze for food \cite{rezaei2021real}. For this problem, we compared D4's performance with the current state-of-the-art solutions including traditional point-process SSMs \cite{truccolo2005point} and GRU-RNNs \cite{chung2014empirical}. In our assessment of different decoder models' performance, we consider mean squared error (MSE), mean absolute error (MAE) \cite{chai2014root}, correlation coefficients (CC), and the 95\% highest posterior density region (HPD) metrics \cite{yao2009bayesian}. The details for the implementation of the experiments including the learning rate, optimization, etc. are discussed in appendix \ref{app_experiment_details}.
\subsection{Langevin dynamics with point-process observations}
Langevin dynamics is used to describe the acceleration of a particle in a liquid. Here we consider one-dimensional harmonic oscillators, with potential function $U(q) = \frac{Kq^2}{2}, q \in \mathbb{R}$ which is a standard test case for Langevin dynamics \cite{leimkuhler2013robust},
\begin{equation}
\label{eq_lungevin}
    dq = M^{-1} p dt, \ \  dp = -\triangledown U(q) dt - \gamma p dt + \sigma M^{0.5}dW
\end{equation}
where $q,p \in \mathbb{R}^{3N}$ are vectors of instantaneous position and momenta, respectively, $W = W(t)$ is a vector of $3N$ independent Wiener processes, $\gamma > 0$ is a free (scalar) parameter, and $M$ is a constant diagonal mass matrix. We simulated 100-dimensional spiking time series using a point process intensity model, see Appendix \ref{app_point_process_obs}. Figure\ref{fig:D4_Langevian}.A shows the simulated spiking data. For the D4, we assume the prediction process is a Gaussian process, $p(x_k|\boldsymbol{y}_k, \boldsymbol{h}_k,\boldsymbol{\Omega})\sim N( \mu_{\boldsymbol{\Omega}}( \boldsymbol{y}_k ,\boldsymbol{h}_k), \sigma_{\boldsymbol{\Omega}}( \boldsymbol{y}_k ,\boldsymbol{h}_k))$. $\mu_{\boldsymbol{\Omega}}( \boldsymbol{y}_k ,\boldsymbol{h}_k)$ and $\sigma_{\boldsymbol{\Omega}}( \boldsymbol{y}_k ,\boldsymbol{h}_k)$ are the mean and standard deviation of the Gaussian predictor which are nonlinear functions of the current, $\boldsymbol{y}_k$, and history of observed spiking data, $\boldsymbol{h}_k$. Given the observed spiking data, Algorithm \ref{alg_D4_learning} simultaneously updates the state transition and prediction process parameters that maximize the $\boldsymbol{Q}$ function given $\boldsymbol{y}_{1:K}$. Figure \ref{fig:D4_Langevian}.C shows the performance of D4, with the optimal hyper-parameter $\lambda$ identified in Figure \ref{fig:D4_Langevian}.B, along with the DDD and DKF. The D4 successfully inferred the latent state trajectory as most of the points in the trajectory are covered by the 95\% HPD of the D4's predictions. The D4 gives higher performance measures, meaning that it gives a better estimation of the underlying states.
\begin{figure}[]
		\centering
		\includegraphics[width = .7\columnwidth]{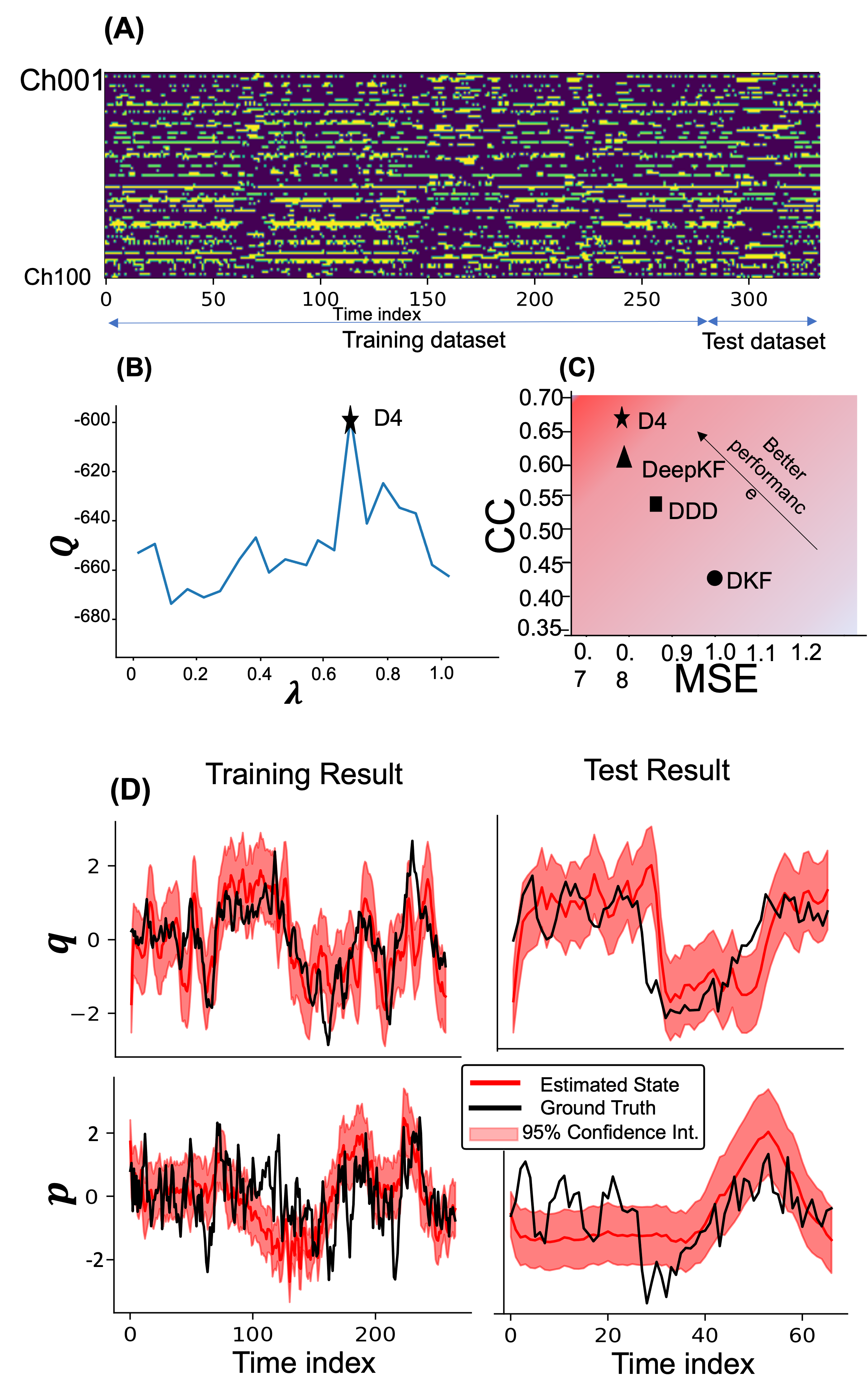}
		\caption{ \centering The D4 inference result for the Langevin system. A) 100 simulated spiking observations based on equations \ref{eq_lungevin}. B) $\lambda$ hyperparameter optimization for equation \ref{eq-objective-final}. C) The D4's performance comparison with DDD \cite{rezaei2022direct} and DKF\cite{burkhart2020discriminative}, Deep-KF\cite{krishnan2015deep}. D) The D4 inferred trajectory for the Langevin latent states.
 }
		\label{fig:D4_Langevian}
	\end{figure}
\subsection{Lorenz Attractor with point-process observations}
\label{Lorenz-section}
Lorenz attractor is a chaotic system \cite{afraimovich1977origin} with its nonlinear dynamics defined by,
\begin{equation}
\label{eq:Lorenz}
    \dot{x_1} = \sigma(x_2 - x_1)+\epsilon_1, \ \  \dot{x_2} = x_1(\rho -x_3)-x_2+\epsilon_2, \ \ \dot{x_3} =x_1 x_2 -\beta x_3+\epsilon_3.
\end{equation}
where the $\{\epsilon_1,\epsilon_2,\epsilon_3\}$ are Gaussian white noises. The model were set to $\{\sigma = 10, \rho = 28, \beta = 8/3\}$ to have a complex trajectory. We simulated $ 100$ spiking data using a point process intensity model, see Appendix\ref{app_point_process_obs}, Figure\ref{fig:D4_Lorenz}.A shows the spiking data.\\
Here, we assume the prediction process is a Gaussian process, $p(x_k|\boldsymbol{y}_k, \boldsymbol{h}_k,\boldsymbol{\Omega})\sim N( \mu_{\boldsymbol{\Omega}}( \boldsymbol{y}_k ,\boldsymbol{h}_k), \sigma_{\boldsymbol{\Omega}}( \boldsymbol{y}_k ,\boldsymbol{h}_k))$, where $\mu_{\boldsymbol{\Omega}}( \boldsymbol{y}_k ,\boldsymbol{h}_k)$ and $\sigma_{\boldsymbol{\Omega}}( \boldsymbol{y}_k ,\boldsymbol{h}_k)$ are the mean and standard deviation of the Gaussian noise, both nonlinear functions of the current, $\boldsymbol{y}_k$, and the history, $\boldsymbol{h}_k$. Figure \ref{fig:D4_Lorenz}.E shows the D4, with the optimal value of $\lambda$ hyper-parameter identified in Figure\ref{fig:D4_Lorenz}.D, and inference performance compared to DDD and DKF models. Similar to the D4's results for Langevin dynamics, shown in Figure \ref{fig:D4_Lorenz}.C, the D4 accurately inferred the latent state trajectories for the Lorenz system with a better performance compared to DDD and DKF.
\subsection{Decoding rat hippocampus data}
 \begin{figure}[]
		\centering
		\includegraphics[width = .8\columnwidth]{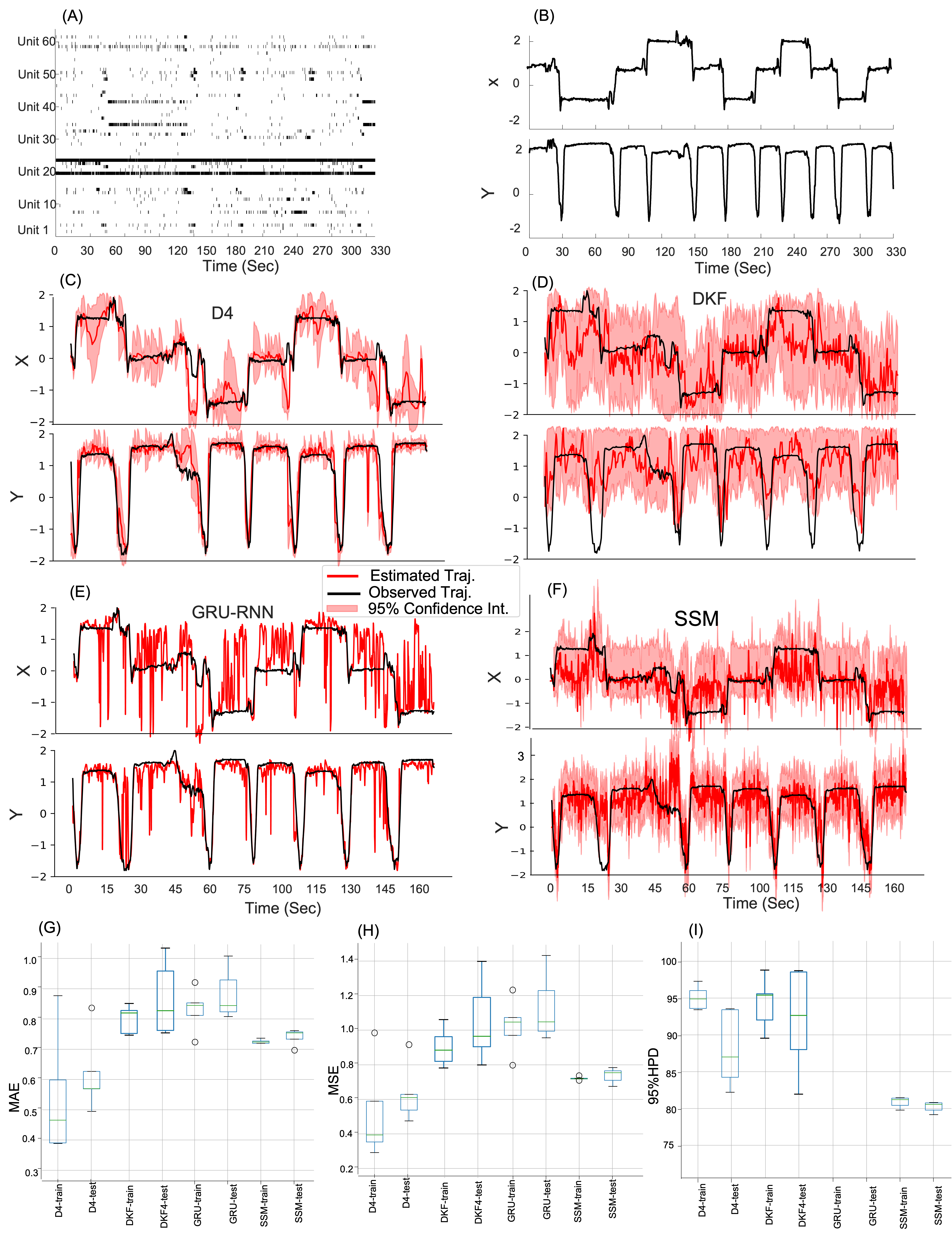}
		\caption{ \centering Decoding result for the rat movement trajectory using D4, DKF, GRU-RNN, and SSM. A) Raster plot of the 62 place cells spiking activity for a 330 seconds period. B) The rat position during the time period shown in A. The rat position is measured using a camera installed over the maze. C-F) D4, DKF, GRU-RNN, and SSM results for the test and training dataset. G-I) MAE, MSE, and 95\% highest posterior density (HPD) performance measures of these models in the rat movement 2D-trajectory decoding.}
		\label{fig:D4_Maze_obs}
	\end{figure}
 In this problem, we seek to decode the 2-D movement trajectory of a rat traversing through a W-shaped maze from the ensemble spiking activity of 62 hippocampal place cells. The neural data are recorded from 62 place cells in the CA1 and CA2 regions of the hippocampus brain area of a rat, more details are available in Appendix\ref{app_neural_data}. Figure\ref{fig:D4_Maze_obs}.A shows the spiking activity of these 62 units, ${\boldsymbol{y}}_{{k}} \in \mathbb{R}^{62}$. Here, the state variable $\{\boldsymbol{x}_k, \boldsymbol{y}_k\}$ represents the rat coordinates in the 2D spaces, see Figure\ref{fig:D4_Maze_obs}.B. For the state process, we use a 2-dimensional random walk with a multivariate normal distribution as the state process. The prediction process is a nonlinear multi-variate regression model for 2D coordinates where the mean and variance are a function of the current spiking activity of 62 cells and their spiking history. Along with the D4, we also build point-process SSM, explained in \cite{truccolo2005point}, DKF\cite{burkhart2020discriminative}, and GRU-RNN \cite{chung2014empirical} models, see Figure \ref{fig:D4_Maze_obs} for decoding results of these models.\\
In this example, unlike the previous examples, we treat the state variables as observed using the 2D trajectories. This means we replace the sampling step for trajectories with the rat's 2D trajectories. Figure\ref{fig:D4_Maze_obs}.G-I shows the performance of different decoding solutions. Among these models, the D4 and DKF are giving a significantly better 95\% HPD compared to the SSM model. Note that this comparison is not possible for GRU-RNN, as its output is deterministic. The D4 model outperforms SSM, GRU-RNN, and DKF in both MAE and MSE performance measures. The D4 reaches higher performance given it optimally combines the information carried by  spiking activity and the state model of the movement at different temporal scales. These results align with the neurons' physiology where their spiking activity is dependent on their previous spikes, a phenomenon not necessarily addressed in SSMs \cite{truccolo2005point}. 

\section{Conclusion}
In this research, we introduced a new modeling framework that extends the utility of SSMs in the analysis of high-dimensional time-series data. We call the solution the Deep Direct Discriminative Decoder, or D4, model. For this model, we demonstrated its solution in both simulated and real datasets, where D4 performs better than RNN and SSM models. The D4 is a variant of SSMs, in which the conditional distribution of the observed signals is replaced by a discriminative process. The D4 inherits attributes of SSM, while it provides a higher expressive power in its prediction of the state variables. In sum, the D4 inherits the advantages of both SSM and DNNs. The D4 can incorporate any information from the history of observed data at different time scales in computing the estimate of this state process. D4 is fundamentally different from SSMs, where the information at only two-time scales: a) fast, which is carried by the observation, and b) slow, defined by the state process, are combined in estimating the state process. We expect the D4 performance to be better than SSM and DNNs, as a D4 without the state process turns to a DNN, and the D4 without the history term will (potentially) correspond to the DKF and/or SSM. Here, we showed that the D4 performance in the simulated data with high-dimensional observation and non-linear state processes preceded the state-of-art solutions including SSM and RNNs. The D4 decoding performance in the neural data is significantly better than SSM and RNNs, showing its applicability in the analysis of complex and high-dimensional time-series data.
\section*{Conflict of interest}
\section*{Acknowledgements}
This work was supported by Dr. Lankarany's NSERC Fund (RGPIN-2020-05868) and MITACS Accelerate (IT19240). A sample of the D4 code is available here:\url{https://github.com/MrRezaeiUofT/Deep_Direct_Discriminative_Decoder-D4-.git}

\printendnotes

\renewcommand{\bibsection}{}
\section*{References}
\bibliography{main}


\appendix
\section{Calculating Q function gradients}
\label{app_gradients}
First we start with calculation of the gradients of the $\boldsymbol{Q}(\boldsymbol{\theta}|\boldsymbol{\theta}^{r})$ with respect to ${\boldsymbol{\Omega}}$ which is defined by
\begin{equation*}
	     \triangledown_{\boldsymbol{\Omega}} \boldsymbol{Q}(\boldsymbol{\theta}|\boldsymbol{\theta}^{r}) = \triangledown_{\boldsymbol{\Omega}} ( \mathbb{E}_K
	[\log \; p({\boldsymbol{x}}_{{0}};\boldsymbol{\omega}_0) + \sum_{k=1}^{K} \log\; p({\boldsymbol{x}}_{{k}}|{\boldsymbol{x}}_{{k-1}};\boldsymbol{\omega}) + \sum_{k=1}^{K} \log \; p({\boldsymbol{x}}_{{k}}|{\boldsymbol{y}}_{{k}},{\boldsymbol{h}}_{{k}};\boldsymbol{\Omega})]
\end{equation*}
\begin{equation*}
    -
	\lambda \sum_{k=1}^{K} \mathbb{KL}(p(\boldsymbol{x}_{k}|\boldsymbol{y}_{1:K},\boldsymbol{\theta}^{r}) \parallel p({\boldsymbol{x}}_{{k}}|{\boldsymbol{h}}_{{k}};\boldsymbol{\Omega})))
\end{equation*}
We can change the order of expectation - $\mathbb{E}_K$- and gradient; thus, we have
\begin{equation*}
	     \triangledown_{\boldsymbol{\Omega}} \boldsymbol{Q}(\boldsymbol{\theta}|\boldsymbol{\theta}^{r}) = \mathbb{E}_K 
	[\triangledown_{\boldsymbol{\Omega}} (\log \; p({\boldsymbol{x}}_{{0}};\boldsymbol{\omega}_0)) + 
	\sum_{k=1}^{K} \triangledown_{\boldsymbol{\Omega}}  ( \log\; p({\boldsymbol{x}}_{{k}}|{\boldsymbol{x}}_{{k-1}};\boldsymbol{\omega})) + \sum_{k=1}^{K} \triangledown_{\boldsymbol{\Omega}}  (\log \; p({\boldsymbol{x}}_{{k}}|{\boldsymbol{y}}_{{k}},{\boldsymbol{h}}_{{k}};\boldsymbol{\Omega}))]
\end{equation*}
\begin{equation*}
    -
	\lambda \sum_{k=1}^{K} \triangledown_{\boldsymbol{\Omega}}  (\mathbb{KL}(p(\boldsymbol{x}_{k}|\boldsymbol{y}_{1:K},\boldsymbol{\theta}^{r}) \parallel p({\boldsymbol{x}}_{{k}}|{\boldsymbol{h}}_{{k}};\boldsymbol{\Omega})))
\end{equation*}
The gradient of the first two term with respect to ${\boldsymbol{\Omega}}$ is zero; as a result, we can rewrite the gradient as
\begin{equation*}
	     \triangledown_{\boldsymbol{\Omega}} \boldsymbol{Q}(\boldsymbol{\theta}|\boldsymbol{\theta}^{r}) = \mathbb{E}_K 
	[ \sum_{k=1}^{K} \triangledown_{\boldsymbol{\Omega}}  (\log \; p({\boldsymbol{x}}_{{k}}|{\boldsymbol{y}}_{{k}},{\boldsymbol{h}}_{{k}};\boldsymbol{\Omega}))]
\end{equation*}
\begin{equation}
\label{app_grad_O_1}
    -
	\lambda \sum_{k=1}^{K} \triangledown_{\boldsymbol{\Omega}}  (\mathbb{KL}(p(\boldsymbol{x}_{k}|\boldsymbol{y}_{1:K},\boldsymbol{\theta}^{r}) \parallel p({\boldsymbol{x}}_{{k}}|{\boldsymbol{h}}_{{k}};\boldsymbol{\Omega})))
\end{equation}
The $\mathbb{KL}(.)$ term can be rewritten as \cite{blei2017variational}
\begin{equation*}
    \mathbb{KL}(p(\boldsymbol{x}_{k}|\boldsymbol{y}_{1:K},\boldsymbol{\theta}^{r}) \parallel p({\boldsymbol{x}}_{{k}}|{\boldsymbol{h}}_{{k}};\boldsymbol{\Omega})) 
\end{equation*}
\begin{equation}
\label{app_grad_O_2}
    = \mathbb{E}_{p(\boldsymbol{x}_{k}|\boldsymbol{y}_{1:K},\boldsymbol{\theta}^{r})}[\log p(\boldsymbol{x}_{k}|\boldsymbol{y}_{1:K},\boldsymbol{\theta}^{r})] - \mathbb{E}_{ p(\boldsymbol{x}_{k}|\boldsymbol{y}_{1:K},\boldsymbol{\theta}^{r})}[\log p({\boldsymbol{x}}_{{k}}|{\boldsymbol{h}}_{{k}};\boldsymbol{\Omega})]
\end{equation}
By replacing equation \ref{app_grad_O_2} inside equation \ref{app_grad_O_1}, we derive
\begin{equation*}
	     \triangledown_{\boldsymbol{\Omega}} \boldsymbol{Q}(\boldsymbol{\theta}|\boldsymbol{\theta}^{r}) = \mathbb{E}_K 
	[ \sum_{k=1}^{K} \triangledown_{\boldsymbol{\Omega}}  (\log \; p({\boldsymbol{x}}_{{k}}|{\boldsymbol{y}}_{{k}},{\boldsymbol{h}}_{{k}};\boldsymbol{\Omega}))]
\end{equation*}
\begin{equation}
\label{app_grad_O_3}
    -
	\lambda \sum_{k=1}^{K} \triangledown_{\boldsymbol{\Omega}}  (\mathbb{E}_{p(\boldsymbol{x}_{k}|\boldsymbol{y}_{1:K},\boldsymbol{\theta}^{r})}[\log p(\boldsymbol{x}_{k}|\boldsymbol{y}_{1:K},\boldsymbol{\theta}^{r})] - \mathbb{E}_{p(\boldsymbol{x}_{k}|\boldsymbol{y}_{1:K},\boldsymbol{\theta}^{r})}[\log p({\boldsymbol{x}}_{{k}}|{\boldsymbol{h}}_{{k}};\boldsymbol{\Omega})])
\end{equation}
The $\mathbb{E}_{p(\boldsymbol{x}_{k}|\boldsymbol{y}_{1:K},\boldsymbol{\theta}^{r})}[p(\boldsymbol{x}_{k}|\boldsymbol{y}_{1:K},\boldsymbol{\theta}^{r})]$ is not dependent on $\boldsymbol{\Omega}$; therefore, the $\triangledown_{\boldsymbol{\Omega}} (\mathbb{E}_{p(\boldsymbol{x}_{k}|\boldsymbol{y}_{1:K},\boldsymbol{\theta}^{r})}[\log p(\boldsymbol{x}_{k}|\boldsymbol{y}_{1:K},\boldsymbol{\theta}^{r})])=0$. Now, we move the gradient inside the expectation term one more time which gives us 
\begin{equation}
\label{app_grad_O_4}
	     \triangledown_{\boldsymbol{\Omega}} \boldsymbol{Q}(\boldsymbol{\theta}|\boldsymbol{\theta}^{r}) = \sum_{k=1}^{K} \mathbb{E}_{ p(\boldsymbol{x}_{k}|\boldsymbol{y}_{1:K},\boldsymbol{\theta}^{r})} 
	[ \triangledown_{\boldsymbol{\Omega}}  (\log \; p({\boldsymbol{x}}_{{k}}|{\boldsymbol{y}}_{{k}},{\boldsymbol{h}}_{{k}};\boldsymbol{\Omega}))
    +
	\lambda  \triangledown_{\boldsymbol{\Omega}} (\log p({\boldsymbol{x}}_{{k}}|{\boldsymbol{h}}_{{k}};\boldsymbol{\Omega}))]
\end{equation}
 Therefore, we can calculate $\triangledown_{\boldsymbol{\Omega} \boldsymbol{Q}(\boldsymbol{\theta}|\boldsymbol{\theta}^{r})}$ by 
	     \begin{equation}
	     \label{app_grad_O_5}
	          \triangledown_{\boldsymbol{\Omega} \boldsymbol{Q}(\boldsymbol{\theta}|\boldsymbol{\theta}^{r})} \simeq \frac{1}{D}\sum_{d=1}^{D} \sum_{k=1}^{K} \triangledown_{\boldsymbol{\Omega}} \log p({\boldsymbol{\hat{x}}}^{(d)}_{{k}}|{\boldsymbol{y}}_{{k}},{\boldsymbol{h}}_{{k}};\boldsymbol{\Omega}) + \lambda \triangledown_{\boldsymbol{\Omega}} \log p({\boldsymbol{\hat{x}}}^{(d)}_{{k}}|{\boldsymbol{h}}_{{k}};\boldsymbol{\Omega})
	     \end{equation}
The equation \ref{app_grad_O_5} is a Monte Carlo estimator of the equation\ref{app_grad_O_5}, where $\{{\boldsymbol{\hat{x}}}^{(d)}_{{k}}\}_{k=1}^{k=K}$ is $d$th sample trajectory from the smoothed posterior with parameter set $\boldsymbol{\theta}^{r}$.\\
Similarly, the gradient  of the $\boldsymbol{Q}(\boldsymbol{\theta}|\boldsymbol{\theta}^{r})$ with respect to ${\boldsymbol{\omega}}$ can be derived  by
\begin{equation}
\label{app_eq_grad_omega}
	     \triangledown_{\boldsymbol{\omega}} \boldsymbol{Q}(\boldsymbol{\theta}|\boldsymbol{\theta}^{r}) \simeq \frac{1}{D}\sum_{d=1}^{D} \sum_{k=1}^{K} \triangledown_{\boldsymbol{\Omega}} \log p({\boldsymbol{\hat{x}}}^{(d)}_{{k}}|{\boldsymbol{\hat{x}}}^{(d)}_{{k-1}};\boldsymbol{\omega})
\end{equation}
Finally, the gradient of the $\boldsymbol{Q}(\boldsymbol{\theta}|\boldsymbol{\theta}^{r})$ with respect to ${\boldsymbol{\omega}_0}$ can be derived  by
\begin{equation}
\label{app_eq_grad_omega0}
	     \triangledown_{\boldsymbol{\omega}_0} \boldsymbol{Q}(\boldsymbol{\theta}|\boldsymbol{\theta}^{r}) \simeq \frac{1}{D}\sum_{d=1}^{D} \triangledown_{\boldsymbol{\Omega}} \log p({\boldsymbol{\hat{x}}}^{(d)}_{{k}}|{\boldsymbol{\hat{x}}}^{(d)}_{{k-1}};\boldsymbol{\omega})
\end{equation}
\section{Performance comparison in 1D simulation data with high-dimensial observations}
\label{app-simulaion-1D}
\begin{figure}[th]
		\centering
		\includegraphics[width = .8\columnwidth]{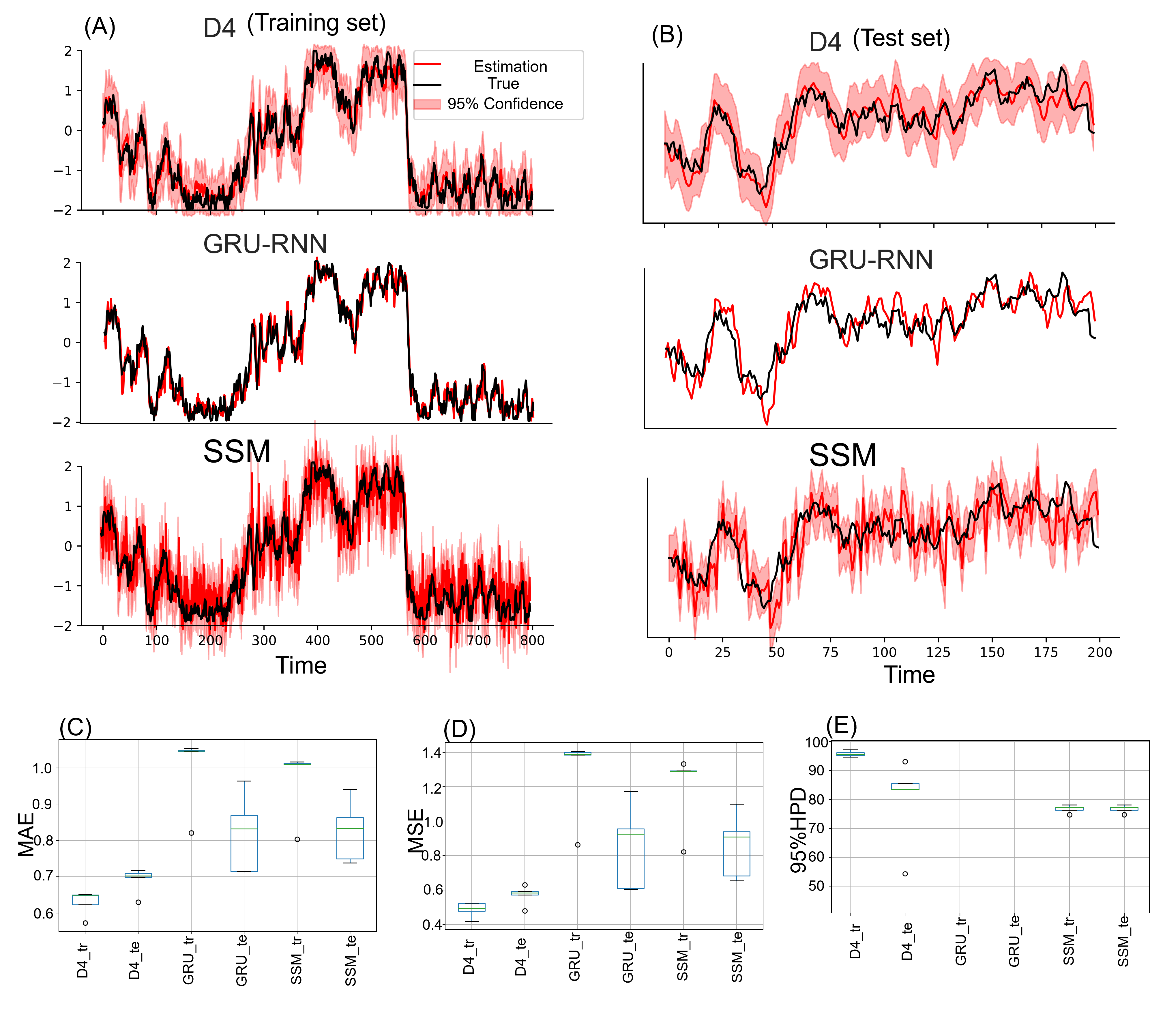}
		\caption{\centering Decoding results for simulation dataset using D4, GRU-RNN, and SSM models. A) Decoding results for the training set. B) Decoding results for the test set. C) MAE. D) MSE. E) 95\% HPD. Note that the GRU-RNN model has a deterministic output; as a result, there is no 95\% HPD measure for the GRU-RNN model.}
		\label{fig:D4_synth_obs}
	\end{figure}
We generate the simulation dataset by assuming the state of interest, $x_k$, is a one-dimensional random walk model which is defined by $P(x_k|x_{k-1})\sim N(ax_{k-1}+b,\sigma_x^2)$, where $a$ is the state transition and $b$ is the bias coefficient for the random walk with a normal additive noise with a standard deviation of $\sigma_x$. Using the state, we then generate a 20-dimensional temporal signal with a conditional distribution defined by
\begin{equation}
    P(\boldsymbol{y}_k|x_k) \sim N(\boldsymbol{g}(x_k, x_{k-1},...,x_{k-l_m}),\boldsymbol{\Sigma}_s)
\end{equation}
where $\boldsymbol{g(.)}$ is a 20-dimensional vector of non-linear functions, like tanh, and cosine, with an argument defined by a subset of state processes at the current and previous time points. The $l_m$ is the maximum length of data points used in $g(.)$ function. For each channel of data, we pick a $l$ uniformly from $0$ to $l_m$ randomly, which is used for data generation. For example, in our simulation, the $g_1$ function corresponding to the first channel of observation $\boldsymbol{g}$, is a $\tanh(.)$ with the argument $x_k+0.8*x_{k-1}+0.6*x_{k-2}+0.4*x_{k-3}+0.2*x_{k-4}$. $\boldsymbol{\Sigma}_s$ is the stationary covariance  matrix with size $20\times20$. $\boldsymbol{\Sigma}_s$’s non-diagonal terms are non-zeros, implying observations across channels are correlated as well. In our simulation, we picked $a = .98$, $b = 0$, and $\sigma_x = 0.1$. We generate the data for 1000 data points. The observation signal in this simulation has complex dynamics while the underlying state is low dimensional; this setting was purposefully picked to demonstrate D4 prediction power and build a clear comparison across models.\\
We tested SSM and GRU-RNN model decoding performance along with the D4. For the training, we assumed the state process is observed. We consider the prediction process is a Gaussian process, $p(x_k|\boldsymbol{y}_k, \boldsymbol{h}_k,\boldsymbol{\Omega})\sim N( \mu_{\boldsymbol{\Omega}}( \boldsymbol{y}_k ,\boldsymbol{h}_k), \sigma_{\boldsymbol{\Omega}}( \boldsymbol{y}_k ,\boldsymbol{h}_k))$. $\mu_{\boldsymbol{\Omega}}( \boldsymbol{y}_k ,\boldsymbol{h}_k)$ and $\sigma_{\boldsymbol{\Omega}}( \boldsymbol{y}_k ,\boldsymbol{h}_k)$ are the mean and standard deviation of the Gaussian predictor which both are nonlinear functions of the current, $\boldsymbol{y}_k$, and the history of observed spiking data, $\boldsymbol{h}_k$. This is because we needed the state for GRU-RNN training, and with the values of the state known, the training of the SSM becomes simple. For SSM training, we can find the state and observation process parameters using the MLE technique, if the state trajectory is known. Figure\ref{fig:D4_synth_obs} shows the D4, GRU-RNN, and SSMs decoding results on this data. The D4 model MAE and MSE measures outperform both SSM and GRU-RNN models. The D4 model also gives even a better 95\% HPD compared to the SSM model.
\section{Point-process observations for Lorenz dynamical system}
\label{app_point_process_obs}
We generate simulated $M$ channels of spiking data using a point process intensity model that is governed by a nonlinear mapping of the state values $\boldsymbol{X}$, where the conditional intensity for each channel of the data is a function of the state process defined by a problem, such as Langevin problem, $\boldsymbol{X}=\{p,q\}$ and Lorenz attractor problem, $\boldsymbol{X}=\{x_1, x_2, x_3\}$, as
\begin{equation}
\label{eq:CIF_PP_S}
    \lambda_j (\boldsymbol{X})= \exp[a_j - \sum_{x_k \in \boldsymbol{X}} \frac{(x_k-\mu_{j,x_k})^2}{2\sigma_{j,x_k}^2}], j=1,..., M
\end{equation}
where $\mu_{j,x_k}$ and $\sigma_{j,x_k}^2$ define the center and width for a hypothetical receptive field model of $x_k$, and $a_j$ is the peak firing rates. The history-dependent terms for $i$th channel is defined by another intensity function as
\begin{equation}
\label{eq:CIF_PP_I}
    \lambda_{j,H}= \sum_{s_n \in S_j}1- \exp( - \frac{(k-s_n)^2}{2\sigma_{j}^2}),
\end{equation}
where $S_j$ is the set containing all the spike times of $j$th channel.
Therefore, the intensity function for $j$th point-process channel, $\hat{\lambda}_j$, is calculated by $\hat{\lambda}_j= \lambda_j * \lambda_{j,H}$. \\
$\mu_{j,x_k}$  are drawn from uniform distributions that cover the domain of the state values, $\sim U(mean(x_k)-2*std(x_k),mean(x_k)+2*std(x_k))$. $\{\sigma_{j,x_k}, \sigma_{i}\}$ are drawn from a uniform distribution, $\sim U(min_{sigma},1/M)$. $a_j$ are drawn from a uniform distribution, $\sim U(min_{firing-rate},max_{firing-rate})$.
\section{Details about the neural data}
\label{app_neural_data}
In this example, we seek to decode the 2D movement trajectory of a rat traversing a W-shaped maze from the ensemble spiking activity of 62 hippocampal place cells \cite{yousefi2019efficient}. The neural data were recorded from an ensemble of place cells in the CA1 and CA2 regions of the hippocampus of a single Long-Evans rat, aged approximately 6 months. The rat was trained to traverse between a home box and the outer arms to receive a liquid reward (condensed milk) at the reward locations. The spiking activity of these 62 units was detected offline by identifying events with peak-to-peak amplitudes above a threshold of 80 uV in at least one of the tetrode channels. The rat’s position was measured using video tracking software and was used for training the models and as the ground truth for the decoded position. We used a 15-minute section of the experiment, with a time resolution of 33 milliseconds, to analyze multiple decoding methods. The first 82\% of the recording (about 12.5 minutes) was used to train the discriminative and state process models and the remaining 18\% (about 2.5 minutes) of data was used to test the model’s decoding
performance.
\section{Implementation details}
In this section, we provide details on the exact setup for each of our experiments. 
\subsection{Langevin dynamics problem}
In this problem for the Gaussian predictor, we considered fully connected neural networks with three layers (hidden size 20), activation function \emph{relu}, dropout rate $50\%$ for both $mu_{\boldsymbol{\Omega}}$ and $\sigma_{\boldsymbol{\Omega}}$. With a greedy search over the hyper-parameter spaces, the number of layers and hidden size are identified. We selected learning of $0.001$ for Adam optimizer with batch size 10 and for 100 iterations of EM. The  Hyperparameters for algorithm \ref{alg_D4_learning} are selected according to a grid search over the hyperparameters as $\{\lambda = 0.5, D=20, L=20 \}$. 
\subsection{Lorenz Attractor problem}
In this problem for the Gaussian predictor, we considered fully connected neural networks with 4 layers (hidden size 20), activation function \emph{relu}, dropout rate $50\%$ for both $mu_{\boldsymbol{\Omega}}$ and $\sigma_{\boldsymbol{\Omega}}$. With a greedy search over the hyper-parameter spaces, the number of layers and hidden size are identified. We selected learning of $0.001$ for Adam optimizer with batch size 10 and for 100 iterations of EM. The  Hyperparameters for algorithm \ref{alg_D4_learning} are selected according to a grid search over the hyperparameters as $\{\lambda = 0.8, D=20, L=25 \}$.
\subsection{Rat hippocampus data}
In this problem for the Gaussian predictor, we considered fully connected neural networks with 4 layers (hidden size 20), activation function \emph{relu}, dropout rate $50\%$ for both $mu_{\boldsymbol{\Omega}}$ and $\sigma_{\boldsymbol{\Omega}}$. We selected learning of $0.001$ for Adam optimizer with batch size 100 and for 100 iterations of EM. The  Hyperparameters for algorithm \ref{alg_D4_learning} are selected according to a grid search over the hyperparameters as $\{\lambda = 0.8, D=50, L=50 \}$.
 \subsection{1-D random walk}
In this problem for the Gaussian predictor, we considered fully connected neural networks with 3 layers (hidden size 10), activation function \emph{relu}, dropout rate $50\%$ for both $mu_{\boldsymbol{\Omega}}$ and $\sigma_{\boldsymbol{\Omega}}$. We selected learning of $0.001$ for Adam optimizer with batch size 20 and for 100 iterations of EM. The  Hyperparameters for algorithm \ref{alg_D4_learning} are selected according to a grid search over the hyperparameters as $\{\lambda = 0.1, D=20, L=10 \}$.

\label{app_experiment_details}
\section{ Deep-KF model performance in Lorenz system}
\begin{figure}[th]
		\centering
		\includegraphics[width = .8\columnwidth]{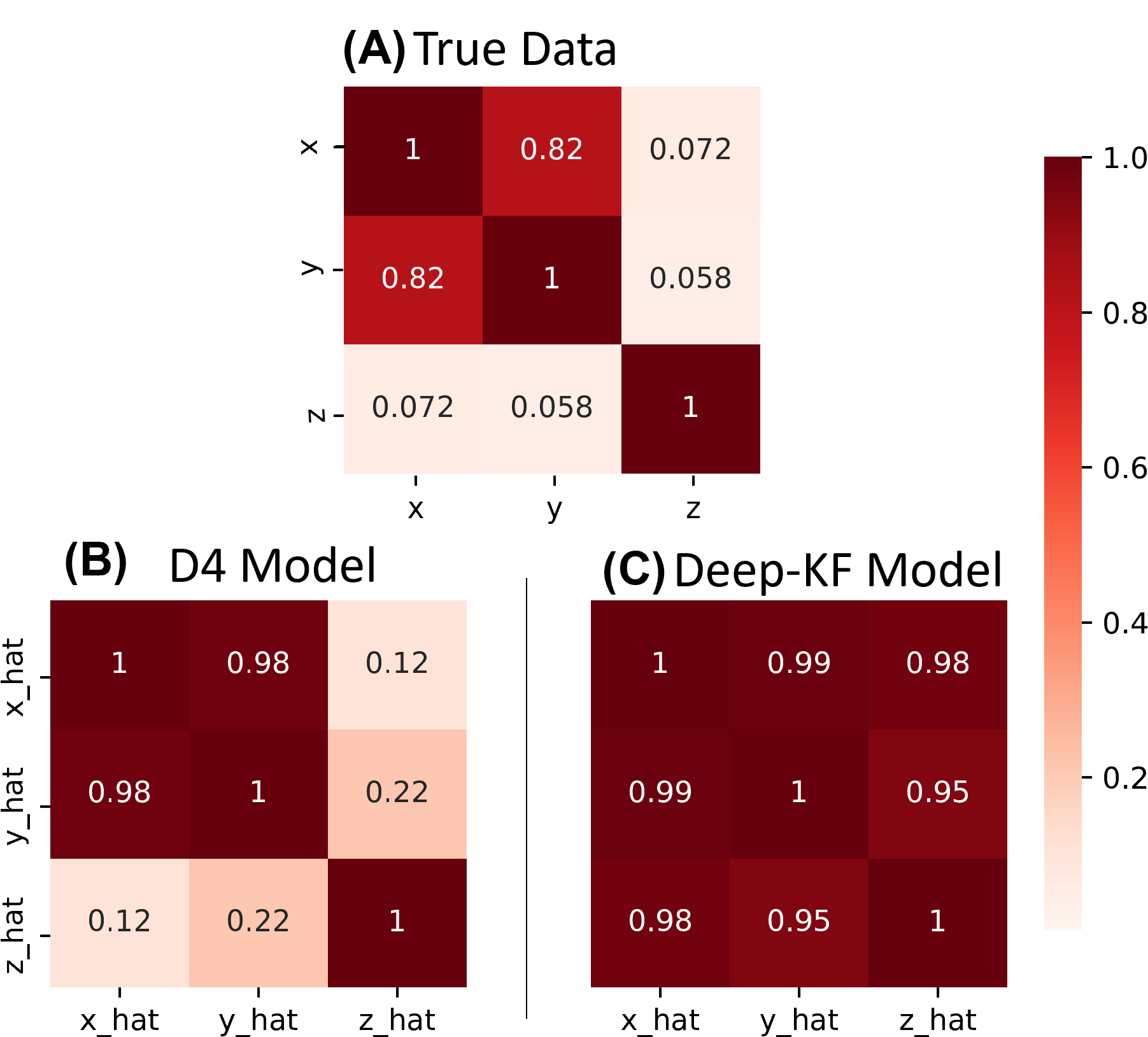}
		\caption{\centering Correlation matrix analysis for A) the true and inferred latent dynamics in the Lorenz problem with B) D4 and C) Deep-KF models.}
		\label{fig:D4_DeepKF_cc}
\end{figure}

\begin{figure}[th]
		\centering
		\includegraphics[width = .8\columnwidth]{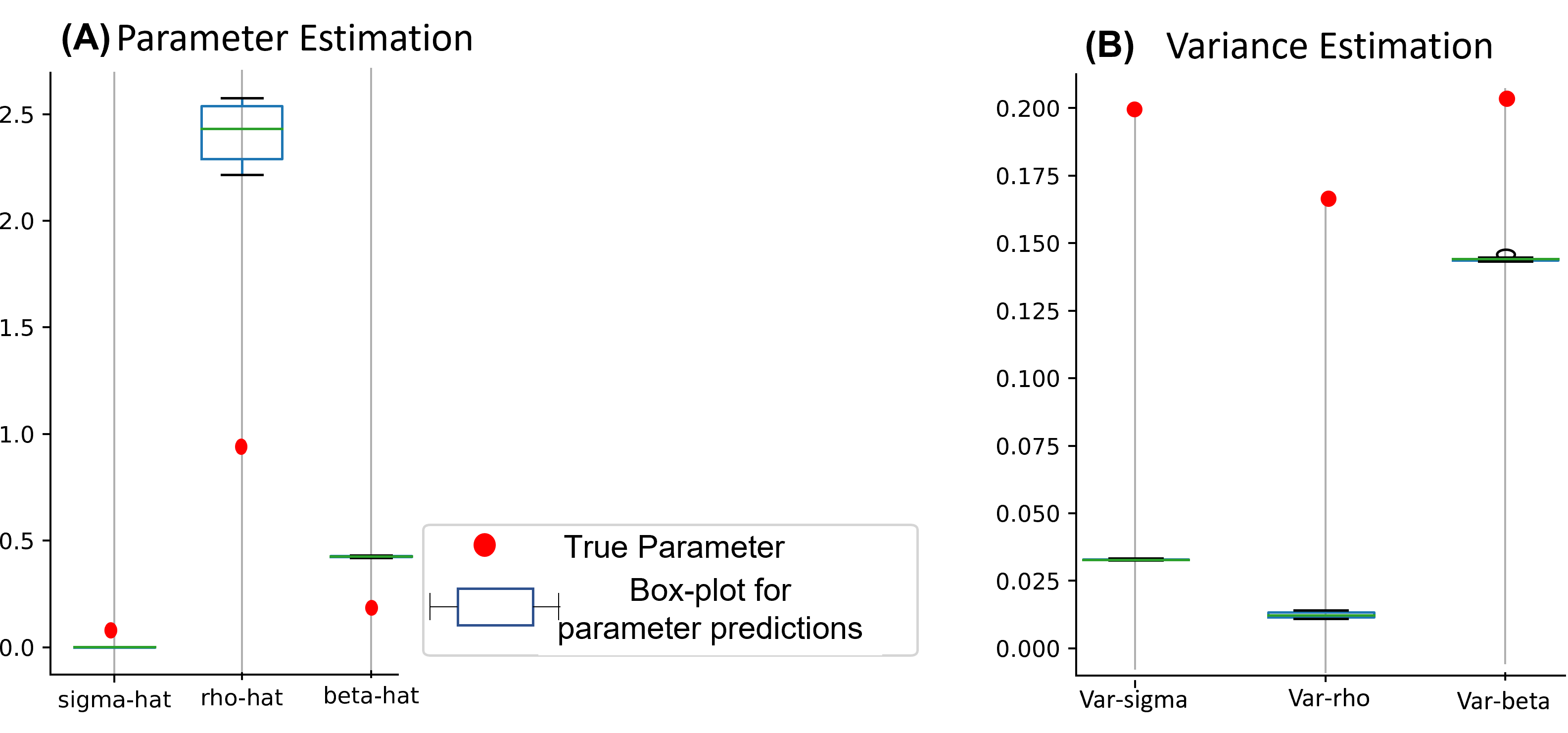}
		\caption{\centering Parameter estimation for inferred latent state with the Deep-KF for the Lorenz attractor. Here we used the inferred state trajectories with 10 different initial points and find the Lorenz system parameters with maximum likelihood. A) Maximum likelihood parameter estimation for the Lorenz system parameters defined in the equation \ref{eq:Lorenz}. The Box plot for each estimated parameter is shown against the true value for the Lorenz system parameter (red  dot) that we used to train the Deep-KF model. B) Variance estimation with the Deep-KF.   }
		\label{fig:D4_DeepKF_ML}
\end{figure}

The deep Kaman filter  (Deep-KF) replaces the state transition and the likelihood models with flexible DNNs whilst it assumes the current observation is conditionally independent of the previous observation given the current state. We previously showed the comparison of D4 and Deep-KF for the Lorenz system in  Figure \ref{fig:D4_Lorenz}. As it has shown in this figure, we reach a better performance measure with the D4 model, especially in the correlation coefficient metric. We believe this improvement is the result of the explicit state process present in the D4, which allows us to correctly track the dynamics of the latent state(s) through noisy observations. On the other hand, Deep-KF has an expressive latent state model which has the tendency to capture observation noise; this will limit the generalization of the model and its ability to draw a proper estimation of the underlying state(s). Figures \ref{fig:D4_DeepKF_cc} and \ref{fig:D4_DeepKF_ML} corroborate our argument here. In Figure \ref{fig:D4_DeepKF_cc}, as it is expected, the correlation of the latent process is preserved under the D4 model. In contrast, for Deep-KF, the correlation of inferred states is completely off from the simulated data pointing out that the Deep-KF state process model infers different dynamics across state variables. Figure \ref{fig:D4_DeepKF_ML} represents the same statement in the context of the state process parameters. When the inferred states of Deep-KF are mapped to the Lorenz equation, we end up with completely different parameter settings with underestimated noise variances.

\end{document}